\documentclass{ifacconf}

\usepackage{amsmath}
\usepackage{amssymb}
\usepackage{graphicx}
\usepackage{natbib}
\usepackage[T1]{fontenc}
\begin{document}
\begin{frontmatter}
\graphicspath{{figures/}}


\title{Towards Effective Physical Reservoir Computing with a Pneumatic Soft Robot}

\thanks[footnoteinfo]{This material is based upon work supported in part by the National Science Foundation under Grant No. 2450946.}

\author[ASU]{Jeevan Hebbal Manjunath}
\author[VT]{Jun Wang}
\author[VT]{Suyi Li}
\author[ASU]{Wenlong Zhang}

\address[ASU]{School of Manufacturing Systems and Networks,
  Arizona State University,
  Mesa, AZ 85212, USA (e-mail: jhebbalm@asu.edu, Wenlong.Zhang@asu.edu)}

\address[VT]{Department of Mechanical Engineering,
  Virginia Tech,
  Blacksburg, VA 24060, USA (e-mail: junw@vt.edu, suyili@vt.edu)}

\begin{abstract}
Physical reservoir computing (PRC) refers to the use of a physical dynamical system as a computational resource for tasks such as state estimation and control, but there has been a lack of formal study of design rules towards more effective design of such physical reservoirs. Using a pneumatic soft arm with a five-pouch sensing column, this work studies how the pouch interconnection topology, robot stiffness, and the number of instrumented sensors affect bending-angle estimation performance. Across 36 matched trials spanning waveform, baseline pressure of the sensing column, and actuation range, all designs are evaluated under the same-time bending-angle estimation benchmark using 0.2~s of pressure history and a fixed ridge estimator. Our analysis of the experimental results leads to three design guidelines. First, independently sealed pouches preserve a much richer observable state than a shared manifold. Second, increasing the baseline pressure of the sensing column makes the pouch responses more redundant and increases estimation error most strongly in the coupled topology. Third, in the sealed topology, two strategically placed sensors already recover most of the attainable benefit, three capture essentially all of it, and additional sensors provide little or no additional value. In summary, the results suggest that topology, stiffness, and number of instrumented sensors should be co-designed for accurate PRC of soft robot states; stronger excitation alone cannot recover the diversity that poor design choices have already removed.
\end{abstract}


\end{frontmatter}

\section{Introduction}
Reservoir computing (RC) uses the transient response of a dynamical system as a computational state and fits only a lightweight readout to map that state to a task variable \citep{jaeger2001echo,maass2002lsm,lukosevicius2009rc}. When the reservoir is a physical body, the framework becomes physical reservoir computing (PRC) \citep{tanaka2019prc}, and soft robots are a natural substrate: compliant structures exhibit nonlinear, history-dependent, and spatially distributed dynamics that are difficult to model explicitly yet rich in information \citep{rus2015soft,nakajima2014softbody,tanaka2019prc}. Prior work has shown that soft-body dynamics can act as a computational resource \citep{nakajima2014softbody,nakajima2015infoproc} and that morphology changes what is available to a downstream controller or decoder \citep{eder2018morph_control}. For soft pneumatic systems, this shifts the sensing problem from a purely instrumentation question to a dynamical-design question: the issue is not only how to measure pressure, but how to shape the pneumatic body so that measured pressures remain informative, diverse, and linearly decodable.

The question matters directly for control: pneumatic soft arms are strongly nonlinear, hysteretic, and configuration-dependent, making state estimation a practical bottleneck in closed-loop operation. Recent pneumatic PRC studies have demonstrated hysteresis compensation and soft-actuator state estimation \citep{shen2025fprc_hysteresis,shen2025gainscheduled_stateest}, and distributed pressure measurements on fabric soft arms can support proprioceptive and exteroceptive perception with minimal training overhead \citep{wang2025ais_prc_fabric}. However, in most existing demonstrations the sensing layout is inherited from the hardware rather than designed systematically: the literature shows that pneumatic PRC can work, but offers little guidance on what sensing morphology to build in the first place.
This work focuses on three coupled design variables that are physically meaningful and experimentally actionable for a multi-pouch pneumatic arm. The first is \emph{topology}: whether the sensing pouches share a manifold or remain independently sealed. The second is \emph{robot stiffness}, controlled here by the baseline pressure $P_0$ of the instrumented sensing column, which governs how deformation redistributes pressure across the body. The third is \emph{sensor count}: how many sensors are instrumented and passed to the estimator. These choices are tightly coupled: topology governs whether local pressure histories survive, stiffness whether they stay distinguishable, and sensor count how much diversity is observed.

Using a five-pouch fabric soft robotic arm, we systematically vary topology, excitation waveform, baseline pressure $P_0$, and actuation range $p_{\max}$ across $2 \times 3 \times 3 \times 2 = 36$ matched trials. This constitutes a controlled benchmark that isolates the three design variables on the same arm while holding the estimator, training split, and feature history fixed; the estimator is intentionally linear throughout, isolating what the hardware makes linearly observable.

Built on this benchmark, the work makes three contributions, one per design axis, each yielding the corresponding design guideline. First, \emph{topology} is the dominant design lever: independently sealed pouches preserve a richer observable pressure state than the shared manifold, yielding an 82\% lower median normalized mean squared error (NMSE) and a $4.5 \times$ larger delay-decoding memory proxy (MC) across matched trials. Second, \emph{robot stiffness} sets the operating regime: increasing baseline pressure drives the pouch responses toward redundancy and degrades estimation most strongly in the coupled manifold, whereas changing actuation range mainly increases bending amplitude without materially changing estimator quality. Third, \emph{sensor count} exhibits strong diminishing returns in the sealed topology: two well-placed sensors recover most of the attainable benefit, and three capture essentially all of it.

\section{Experimental Platform}

\subsection{Soft Arm and Measured Variables}
We use the fabric-based pneumatic soft arm from \citep{qiao2024nonlinear}, built on the actuator design of \citep{nguyen2020fabric}, as our experimental platform. The arm comprises four segments; Segments~2--4 are actively driven by pressure regulators commanded via MCP4725 12-bit DACs on an Arduino, forming the input excitation $\mathbf{u}(t) \in \mathbb{R}^{3}$ to the reservoir.

Segment~1 is the instrumented sensing column and contains five pneumatic pouches, each equipped with an analog pressure transducer (0--30\,PSI range). The five transducer voltages are digitized by two ADS1115 16-bit ADCs (I\textsuperscript{2}C, 4-channel each), with four sensors read on the first ADC and one on the second. Pouch numbering follows the arm from distal to proximal: pouch~1 is closest to the end effector, and pouch~5 is closest to the base. The measured pouch-pressure vector
\begin{equation}
\mathbf{S}(t) = [s_1(t),\ldots,s_5(t)]^{\top}
\end{equation}
serves as the reservoir state. Commands and sensor readings are exchanged with the host PC at 100~Hz.

Ground-truth arm bending is recorded by OptiTrack motion capture using three rigid bodies mounted along the structure. Rigid body~1 is attached to the fixed top reference, rigid body~2 is attached near the beginning of the robotic arm, and rigid body~3 is attached at the end effector. The reported bending angle is computed from the relative orientation between rigid bodies~1 and~3, while rigid body~2 provides an intermediate marker along the arm. Let $\mathbf{q}_1(t)$ and $\mathbf{q}_3(t)$ denote the unit quaternions of rigid bodies~1 and~3, and let $\otimes$ denote quaternion multiplication. The relative rotation quaternion is
\begin{equation}
\mathbf{q}_{\mathrm{rel}}(t) = \mathbf{q}_1^{-1}(t) \otimes \mathbf{q}_3(t),
\end{equation}
and the bending-angle magnitude is
\begin{equation}\label{eq:theta}
\theta(t) = \frac{180}{\pi} \, 2\cos^{-1}\!\left(\left|\mathbf{q}_{\mathrm{rel},w}(t)\right|\right),
\end{equation}
where $\mathbf{q}_{\mathrm{rel},w}$ is the scalar component of $\mathbf{q}_{\mathrm{rel}}$. Pressure and motion-capture data are synchronized at 100~Hz.

\begin{figure}
\begin{center}
\includegraphics[width=8.4cm]{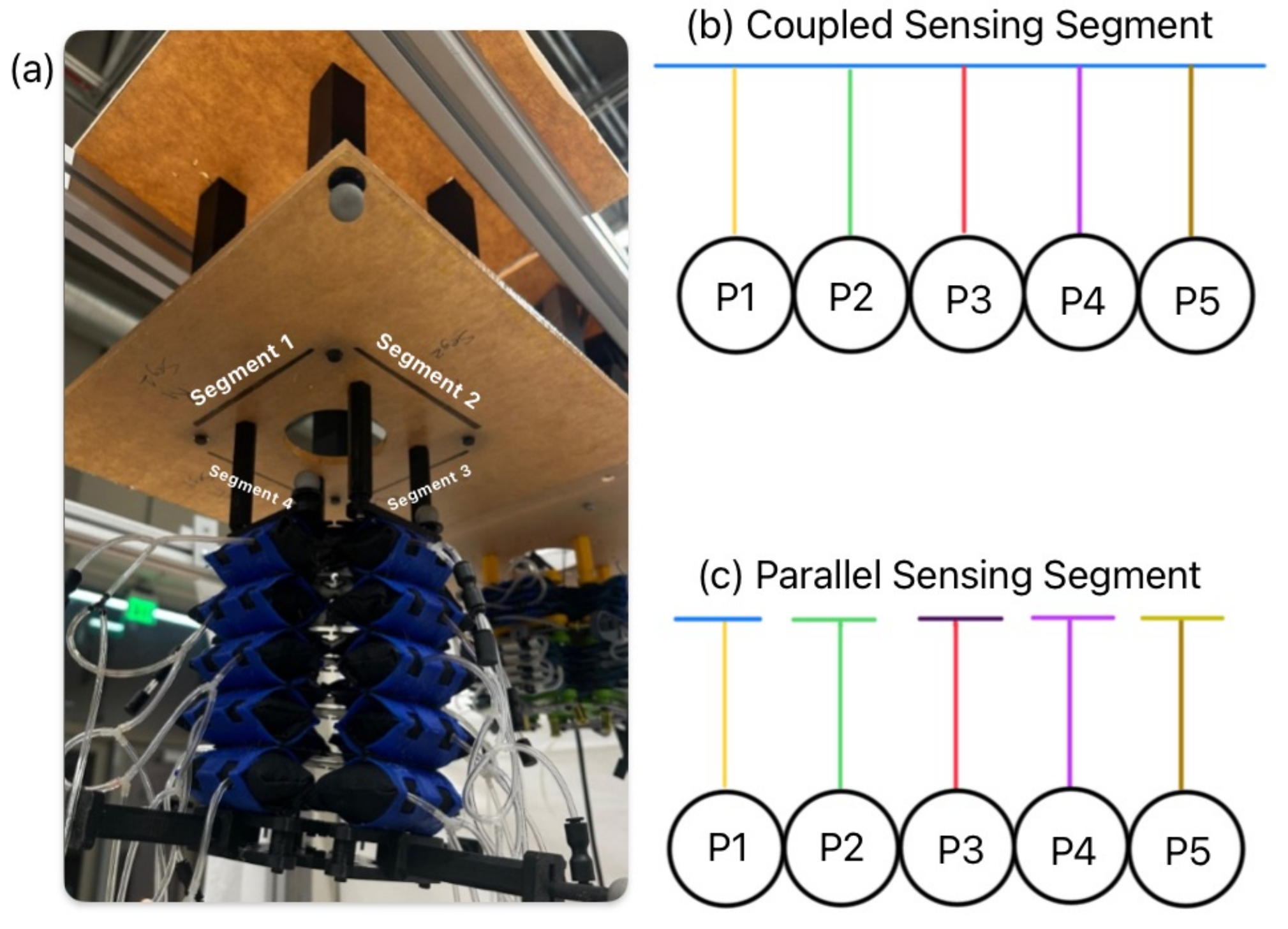}
\caption{Experimental platform and pneumatic sensing topologies. (a) Multi-segment pneumatic arm with the instrumented sensing column (Segment~1). Pouch numbering on Segment~1 runs from distal to proximal: P1 is closest to the end effector, and P5 is closest to the base. Three OptiTrack rigid bodies are mounted at the fixed top reference, near the arm base, and at the end effector. (b) Coupled topology: P1--P5 share a pneumatic manifold, allowing pressure equalization. (c) Independently sealed (parallel) topology: P1--P5 are sealed separately, preventing air exchange and preserving local signal diversity.}
\label{fig:system}
\end{center}
\end{figure}

\subsection{Design Axes and Experimental Protocol}\label{sec:protocol}

The study varies all three design axes explicitly. For \emph{topology}, we test two limiting cases for Segment~1. In the \emph{coupled} configuration, the five pouches are interconnected through a shared manifold: the column is inflated to the baseline pressure $P_0$ through a single common supply line, which is then closed, so during operation air redistributes freely among all five pouches. This strengthens a coherent global pressure mode and suppresses local differences among pouches. In the \emph{independently sealed} or \emph{parallel} configuration, each pouch is inflated to the same baseline pressure $P_0$ through its own line and then sealed individually, so no air exchange occurs between pouches during operation, and each pouch responds primarily to local strain.

For \emph{robot stiffness}, we vary the baseline pressure of Segment~1 over $P_0 \in \{1,2,3\}$~PSI. We report $P_0$ in pressure units because pressure is the experimentally imposed variable, but interpret it physically. Changing $P_0$ changes how the arm resists bending and redistributes strain; $P_0$ serves throughout as the experimental knob for robot stiffness. For \emph{sensor count}, we later evaluate fixed subsets of $m \in \{1,\ldots,5\}$ instrumented sensors.

The final experimental factor is \emph{actuation range}, set by the maximum commanded pressure $p_{\max} \in \{5,10\}$~PSI$\,$ for the driven segments. Increasing $p_{\max}$ enlarges the commanded motion envelope, separating amplitude effects from topology and stiffness effects. The complete study therefore spans $2 \times 3 \times 3 \times 2 = 36$ matched trials: two topologies, three excitation waveforms, three robot-stiffness settings $P_0$, and two actuation maxima $p_{\max}$. Every trial lasts 200~s; the first and last 10~s are discarded to remove start-up and shut-down transients. All the data follow a chronological 70/30 train/test split.

The commanded pressure profiles applied to Segments~2--4 are listed in Table~\ref{tb:actuation}. The sinusoidal primitive is
\begin{equation}\label{eq:sin}
  s(t,\phi) = \frac{p_{\max}}{2}\bigl(1+\sin(2\pi f t + \phi)\bigr),
\end{equation}
with $f = 0.1$~Hz. The triangular primitive is
\begin{equation}\label{eq:tri}
  \mathrm{tri}(\tau;\phi)
    = p_{\max}\left[\min\!\Bigl(\frac{3\tau'}{T},\; 2-\frac{3\tau'}{T}\Bigr)\right]^+,
\end{equation}
where $\tau' = (\tau - \phi T/2\pi) \bmod T$, $\tau = t \bmod T$, $T = 10$~s, and $[\,\cdot\,]^+ = \max(0,\cdot)$. The \emph{axial} primitive drives only Segment~3 and primarily excites one bending direction. The \emph{circular} primitive is the three-phase set $\{s(t,0), s(t,2\pi/3), s(t,4\pi/3)\}$ applied to Segments~2--4; the rotating phase pattern drives the arm through an approximately circular end-effector orbit and is therefore referred to as the circular primitive. The \emph{triangular} primitive uses the same $120^{\circ}$ phase offsets but replaces the sinusoid with piecewise-linear ramps, producing sharper transitions in the commanded motion.

\begin{table}[htbp]
\centering
\caption{Desired actuation pressure profiles $p_{d,i}(t)$ for Segments~2--4.}
\label{tb:actuation}
\setlength{\tabcolsep}{4pt}
\renewcommand{\arraystretch}{1.25}
\begin{tabular}{lcccc}
\hline
Waveform & $p_{d,2}(t)$ & $p_{d,3}(t)$ & $p_{d,4}(t)$ & Trajectory \\
\hline
Axial      & $0$ & $s(t,0)$ & $0$ & \includegraphics[width=0.85cm]{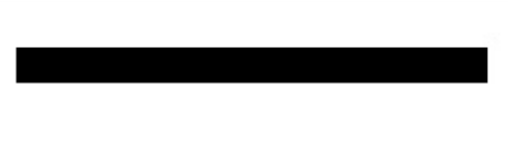} \\
Circular   & $s(t,0)$ & $s(t,2\pi/3)$ & $s(t,4\pi/3)$ & \includegraphics[width=0.85cm]{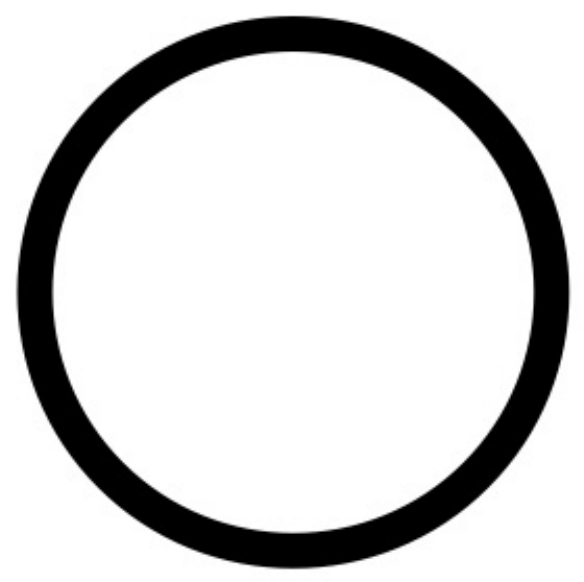} \\
Triangular & $\mathrm{tri}(\tau;2\pi/3)$ & $\mathrm{tri}(\tau;0)$ & $\mathrm{tri}(\tau;4\pi/3)$ & \includegraphics[width=0.85cm]{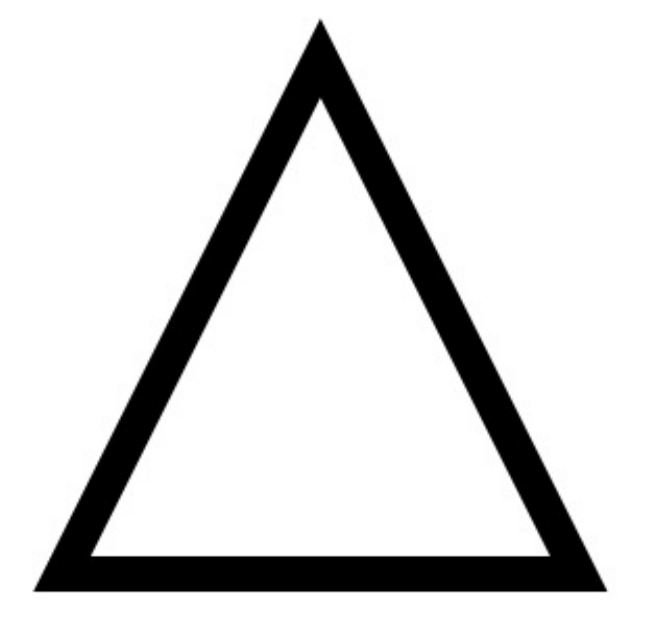} \\
\hline
\end{tabular}
\end{table}

\section{Methods}\label{sec:methods}

\subsection{Fixed Estimator Pipeline}

The estimator exists to compare sensing designs, not to optimize a learning model: it is intentionally observer-like, asking what the sensing layout makes linearly decodable, so its architecture, preprocessing, regularization, and train/test split are identical across all trials. The commanded pressure vector $\mathbf{u}(t) \in \mathbb{R}^3$ applied to Segments~2--4 drives the reservoir dynamics, whereas the estimator itself receives only the measured pouch-pressure state $\mathbf{S}(t) \in \mathbb{R}^5$. To include short-term history, we use a tapped-delay embedding,
\begin{equation}\label{eq:tdl}
\mathbf{x}(t) = [\mathbf{S}(t),\mathbf{S}(t-1),\ldots,\mathbf{S}(t-n_\tau+1)] \in \mathbb{R}^{5n_\tau},
\end{equation}
with $n_\tau = 20$, yielding a 100-dimensional feature vector corresponding to 0.2~s of pressure history at 100~Hz. When only $m < 5$ pouches are instrumented (Sec.~\ref{sec:ablation}), the embedding stacks only the $m$ selected signals, so $\mathbf{x}(t) \in \mathbb{R}^{m n_\tau}$; for example, $m = 3$ yields a 60-dimensional feature vector.

The linear estimator is
\begin{equation}\label{eq:readout}
\hat{\theta}(t) = \mathbf{w}^{\top}\mathbf{x}(t) + w_0,
\end{equation}
where $\mathbf{w} \in \mathbb{R}^{5 n_\tau}$ ($\mathbb{R}^{m n_\tau}$ for sensor subsets) collects the readout weights and $w_0$ is a bias, both fitted on the training split. The readout thus estimates the current bending angle from current and recent pressure history, matching the intended use of the reservoir as a proprioceptive estimator for feedback. Ridge regression with $\alpha = 0.01$ is used after feature standardization; $\alpha$ was selected by a coarse training-set sweep and held fixed for the entire study.

\subsection{Evaluation Metrics}\label{sec:metrics}

No single metric fully characterizes reservoir quality, so we report one task metric, one short-term-memory proxy, and several signal-diversity descriptors. Unless noted, estimation metrics use the held-out test split and signal statistics use the post-transient trial data.

\subsubsection{Estimation error.}
We quantify same-time bending-angle estimation accuracy with normalized mean squared error (NMSE). Let $\theta(t)$ denote the ground-truth test-set angle. Let $\hat{\theta}(t)$ denote the estimate produced by Eq.~\eqref{eq:readout}, and let $\bar{\theta}$ denote the test-set mean angle. Then
\begin{equation}\label{eq:nmse}
\mathrm{NMSE} = \frac{\sum_t \left(\theta(t) - \hat{\theta}(t)\right)^2}{\sum_t \left(\theta(t) - \bar{\theta}\right)^2}.
\end{equation}
NMSE $< 1$ indicates performance better than predicting the test-set mean, and lower values are better. Because topology and stiffness also change $\theta(t)$ itself, absolute error would conflate estimator quality with motion amplitude; NMSE instead scores each trial relative to its own motion variance. As an explicit control, doubling $p_{\max}$ significantly increases bending variability yet leaves NMSE statistically unchanged (Sec.~\ref{sec:regime}), so the reported effects are not amplitude artifacts.

\subsubsection{Delay-decoding memory proxy.}
Rather than the full information-processing capacity of \citet{dambre2012ipc}, we use a delay-decoding memory proxy asking how much past actuation remains linearly recoverable from the current pouch state \citep{lukosevicius2012practical}. For each delay $k \in \{1,\ldots,K\}$, we train a linear readout to reconstruct the past input $\mathbf{u}(t-k)$ from the current state $\mathbf{S}(t)$. Reconstruction quality is measured by the coefficient of determination $R^2$ on held-out data. The average is over the driven Segment~3 input for the axial waveform and over the three driven segments otherwise; denoting it by $\bar{R}^2(k)$, the memory score is

\begin{equation}\label{eq:mc}
\mathrm{MC} = \sum_{k=1}^{K} \max\bigl(0,\bar{R}^2(k)\bigr), \qquad K = 40.
\end{equation}
The upper limit $K = 40$ corresponds to 0.4~s of input history at 100~Hz, which exceeds the 0.2~s tapped-delay window and spans the range over which delayed inputs remain meaningfully decodable under the tested excitation bandwidth. We retain the abbreviation MC below for brevity, but interpret it as a comparative memory proxy rather than a complete capacity analysis.

\subsubsection{Signal diversity and agreement.}
To quantify redundancy among pouches, we use several complementary statistics. The mean inter-pouch Pearson correlation is
\begin{equation}
\bar{r} = \frac{1}{\binom{5}{2}}\sum_{i<j} r_{ij},
\end{equation}
where $r_{ij}$ is the Pearson correlation between z-scored signals of pouches $i$ and $j$. Z-scoring removes offset and scale, so $\bar{r}$ reflects waveform similarity only.

The mean concordance correlation coefficient (CCC) is
\begin{equation}
\bar{\rho}_c = \frac{1}{\binom{5}{2}}\sum_{i<j}
\frac{2\sigma_{ij}}{\sigma_i^2 + \sigma_j^2 + (\mu_i - \mu_j)^2},
\end{equation}
where $\mu_i$ and $\sigma_i^2$ are the mean and variance of pouch $i$, and $\sigma_{ij}$ is the covariance between pouches $i$ and $j$. Unlike Pearson correlation, CCC penalizes mean and amplitude mismatch, so it indicates whether two pouches behave as near-duplicate sensors rather than merely correlated ones \citep{lin1989ccc}.

We additionally report the variance explained by the first principal component (PC1, \%) from principal component analysis (PCA), and the participation ratio

\begin{equation}
\mathrm{PR} = \frac{\left(\sum_i \lambda_i\right)^2}{\sum_i \lambda_i^2},
\end{equation}
where $\lambda_i$ are the eigenvalues of the pouch-signal covariance matrix. PR measures effective dimensionality, from 1 (a single dominant mode) to 5 (five equally contributing modes).
Finally, mean sensitivity --- the average absolute slope of a linear fit from $\theta$ to each pouch pressure, in PSI/deg --- is reported as a secondary descriptor of transduction strength, distinguishing stronger single-sensor pressure--angle coupling from richer multi-sensor state diversity.
\section{Results}\label{sec:results}

\subsection{Topology sets reservoir-state richness}

\begin{figure}
\begin{center}
\includegraphics[width=8.4cm]{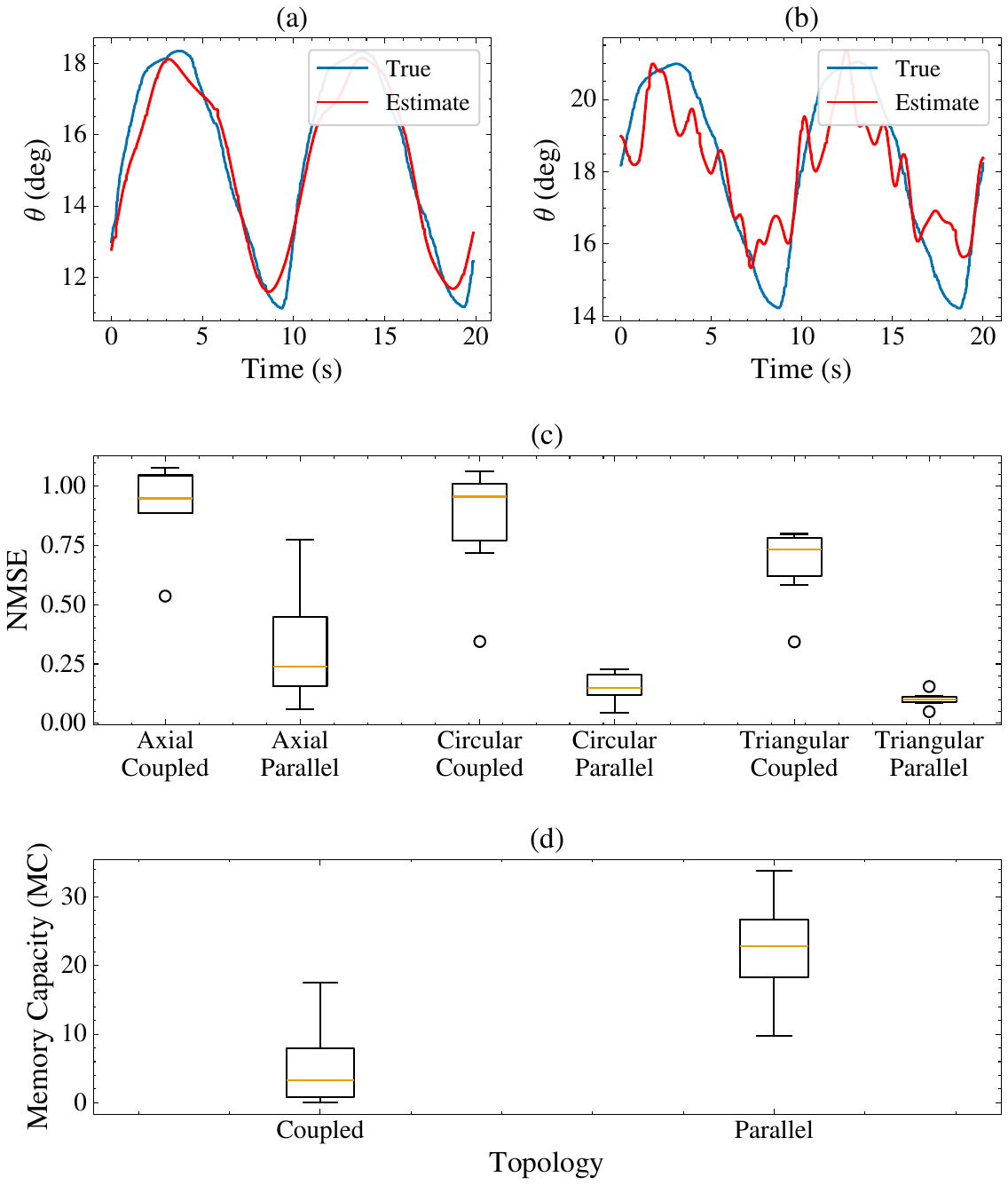}
\caption{Bending-angle estimation performance. (a)--(b) Representative true/estimated traces for matched triangular trials with $P_0 = 1$~PSI and $p_{\max} = 5$~PSI in the sealed and coupled topologies. (c) NMSE across waveforms and topologies (boxes: interquartile range (IQR); center line: median). (d) Delay-decoding memory proxy MC across trials.}
\label{fig:performance}
\end{center}
\end{figure}

Table~\ref{tb:summary} and Fig.~\ref{fig:performance} show the central result: the independently sealed topology yields substantially lower same-time estimation error and much higher short-term memory than the shared manifold. Median NMSE falls from 0.842 to 0.148, an 82\% reduction (paired Wilcoxon signed-rank, two-sided $p = 7.6{\times}10^{-6}$), while mean MC rises from $5.1 \pm 5.1$ to $22.9 \pm 6.5$. The advantage holds across all three excitation families: median NMSE (coupled/sealed) is 0.949/0.238 (axial), 0.956/0.148 (circular), and 0.732/0.100 (triangular). Within a topology, waveform-only differences do not reach significance (Kruskal--Wallis $p = 0.109$ coupled, $p = 0.085$ sealed), so the topology effect is not driven by one specific motion primitive.

\begin{table}[htbp]
\begin{center}
\caption{Aggregate comparison across the 36 trials (18 per topology).}
\label{tb:summary}
\setlength{\tabcolsep}{4pt}
\begin{tabular}{p{3.9cm}cc}
\hline
Metric & Coupled & Sealed \\
\hline
Median estimation NMSE ($\downarrow$) & 0.842 & 0.148 \\
Mean MC ($K{=}40$, $\uparrow$) & $5.1 \pm 5.1$ & $22.9 \pm 6.5$ \\
Pearson corr. $\bar{r}$ ($\downarrow$) & $0.992 \pm 0.005$ & $0.931 \pm 0.055$ \\
CCC $\bar{\rho}_c$ ($\downarrow$) & $0.902 \pm 0.060$ & $0.112 \pm 0.060$ \\
Sensitivity (PSI/deg, $\uparrow$) & $0.0072 \pm 0.0012$ & $0.0044 \pm 0.0015$ \\
\hline
\end{tabular}
\end{center}
\end{table}

\begin{figure}
\begin{center}
\includegraphics[width=0.9\linewidth]{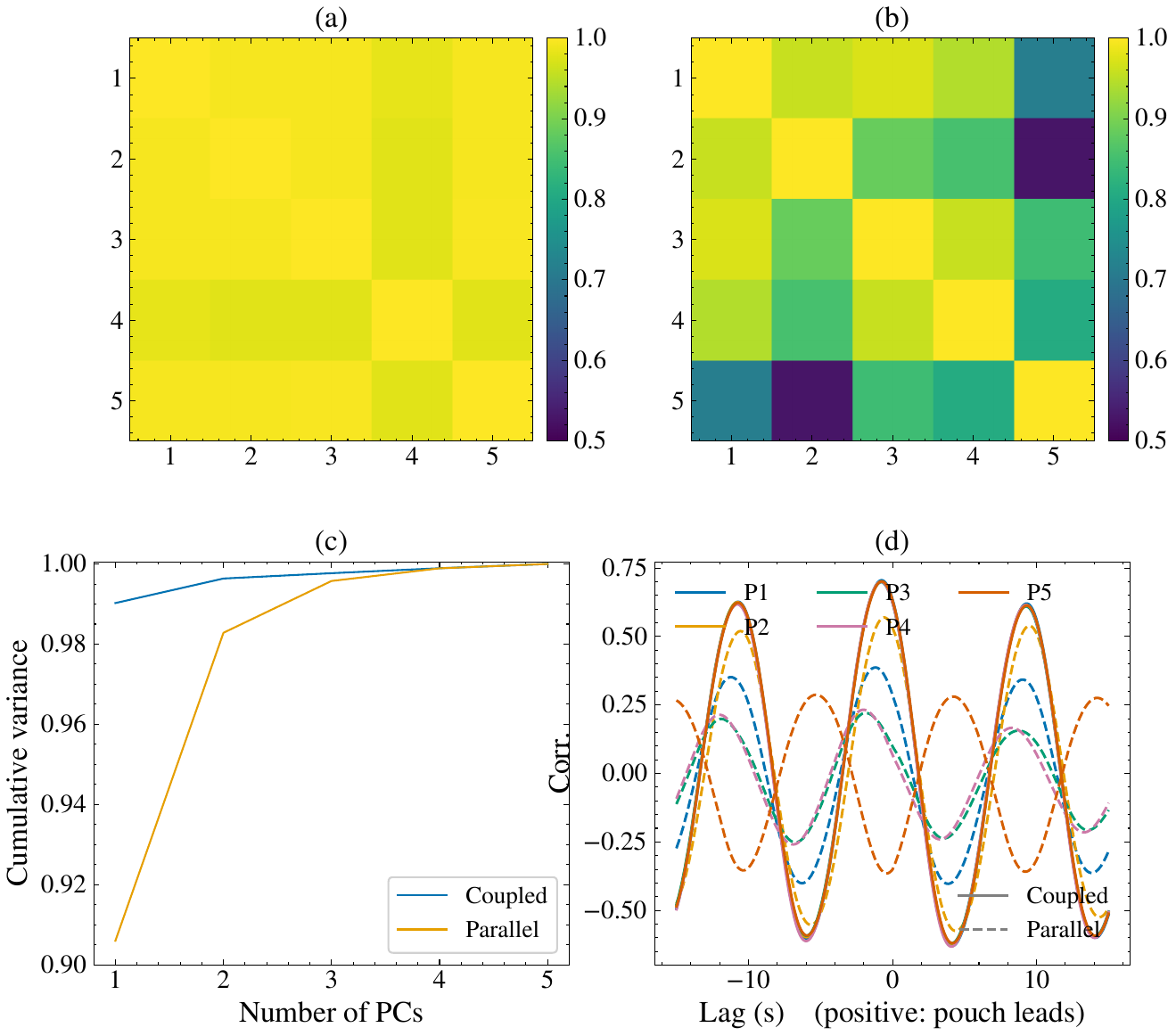}
\caption{Representative diversity trial: triangular excitation, $P_0 = 1$~PSI, actuation range 0--10~PSI. (a)--(b) Inter-pouch correlation matrices. (c) Cumulative PCA variance. (d) Cross-correlation between each pouch and $\theta$.}
\label{fig:diversity}
\end{center}
\end{figure}

The mechanism behind this difference is state diversity rather than raw signal amplitude. In the coupled manifold, air exchange forces the pouches toward an almost single-mode response: the correlation matrix in Fig.~\ref{fig:diversity}(a) is nearly saturated, and PC1 captures about 99.0\% of the variance in the representative coupled trial (Fig.~\ref{fig:diversity}(c)). The 100-dimensional tapped-delay state then contains many delayed copies of essentially the same pressure trajectory. In the sealed topology, inter-pouch agreement is lower, lag profiles relative to $\theta$ are more heterogeneous (Fig.~\ref{fig:diversity}(d)), and the estimator receives more complementary temporal structure. Across all 36 trials, mean inter-pouch correlation tracks worse NMSE (Spearman $\rho = 0.819$, $p = 1.0{\times}10^{-9}$), while participation ratio tracks better NMSE ($\rho = -0.772$, $p = 3.5{\times}10^{-8}$). 

The coupled manifold is not uniformly inferior: it produces stronger sensitivity (0.0072 vs.\ 0.0044~PSI/deg), which may favor plumbing-constrained designs.

\subsection{Robot stiffness and actuation range shape the operating regime}\label{sec:regime}
\begin{figure}
\begin{center}
\includegraphics[width=8.4cm]{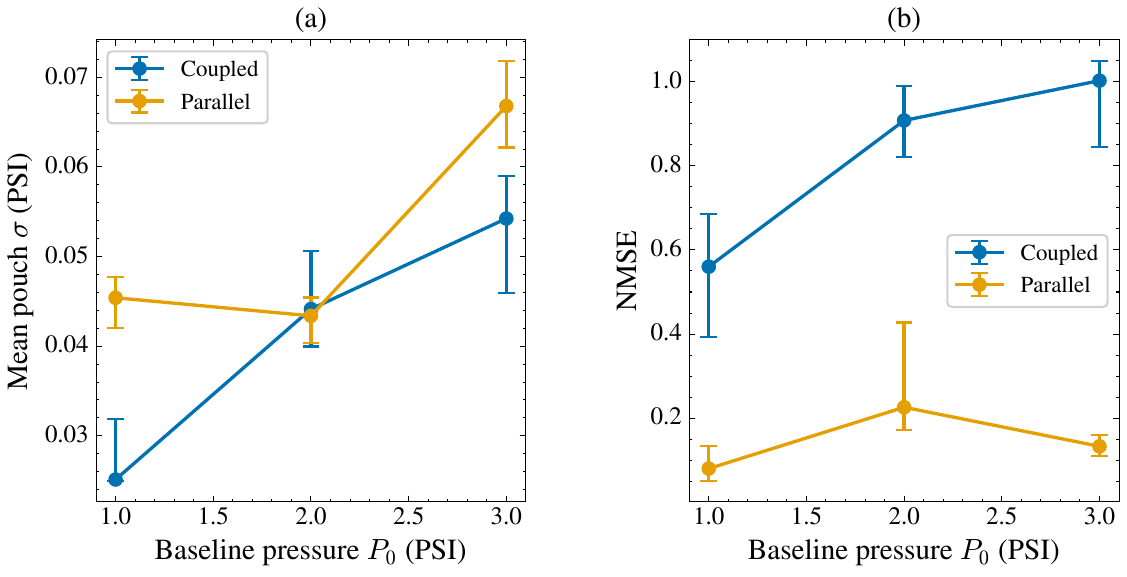}
\caption{Operating regime vs.\ robot stiffness (baseline pressure $P_0$). (a) Mean pouch pressure standard deviation. (b) Estimation NMSE.}\label{fig:regime}
\end{center}
\end{figure}

Baseline pressure $P_0$ reshapes the reservoir primarily through redundancy rather than signal amplitude. Although mean pouch pressure standard deviation increases with $P_0$ in the coupled topology (Fig.~\ref{fig:regime}(a)), estimator performance worsens, ruling out the hypothesis that larger pressure excursions alone improve state quality.

The effect is most pronounced in the coupled topology, where mean inter-pouch correlation rises from 0.986 to 0.995, mean MC collapses from 10.6 to 1.1, and median NMSE deteriorates from 0.560 to 1.002 as $P_0$ increases from 1 to 3~PSI. The sealed topology exhibits the same trend but with greater robustness: median NMSE follows a non-monotonic pattern (0.081, 0.226, 0.133 at 1, 2, 3~PSI), reflecting a competing trade-off between stronger local pressure transduction and increasing inter-pouch redundancy at higher stiffness.

Actuation range has a structurally different effect. Increasing $p_{\max}$ from 5 to 10~PSI produces a significant increase in bending variability across both topologies (paired Wilcoxon: $p = 0.004$ coupled, $p = 0.027$ sealed). However, the corresponding change in NMSE is not significant within either topology ($p = 0.652$ coupled, $p = 0.426$ sealed).

\subsection{Instrumented-sensor count limits observed diversity}\label{sec:ablation}
Throughout this section, $m \in \{1,\ldots,5\}$ denotes the \emph{instrumented-sensor count} (how many pouches the estimator reads), while P1--P5 label \emph{individual pouches} (P1 distal, P5 proximal; Fig.~\ref{fig:system}). For each count $m$, all $\binom{5}{m}$ fixed pouch subsets are exhaustively evaluated in the sealed topology, retaining the subset with the lowest median NMSE across the 18 sealed trials; the coupled topology is evaluated on those same subsets as a baseline reference.

\begin{figure}
\begin{center}
\includegraphics[width=8.4cm]{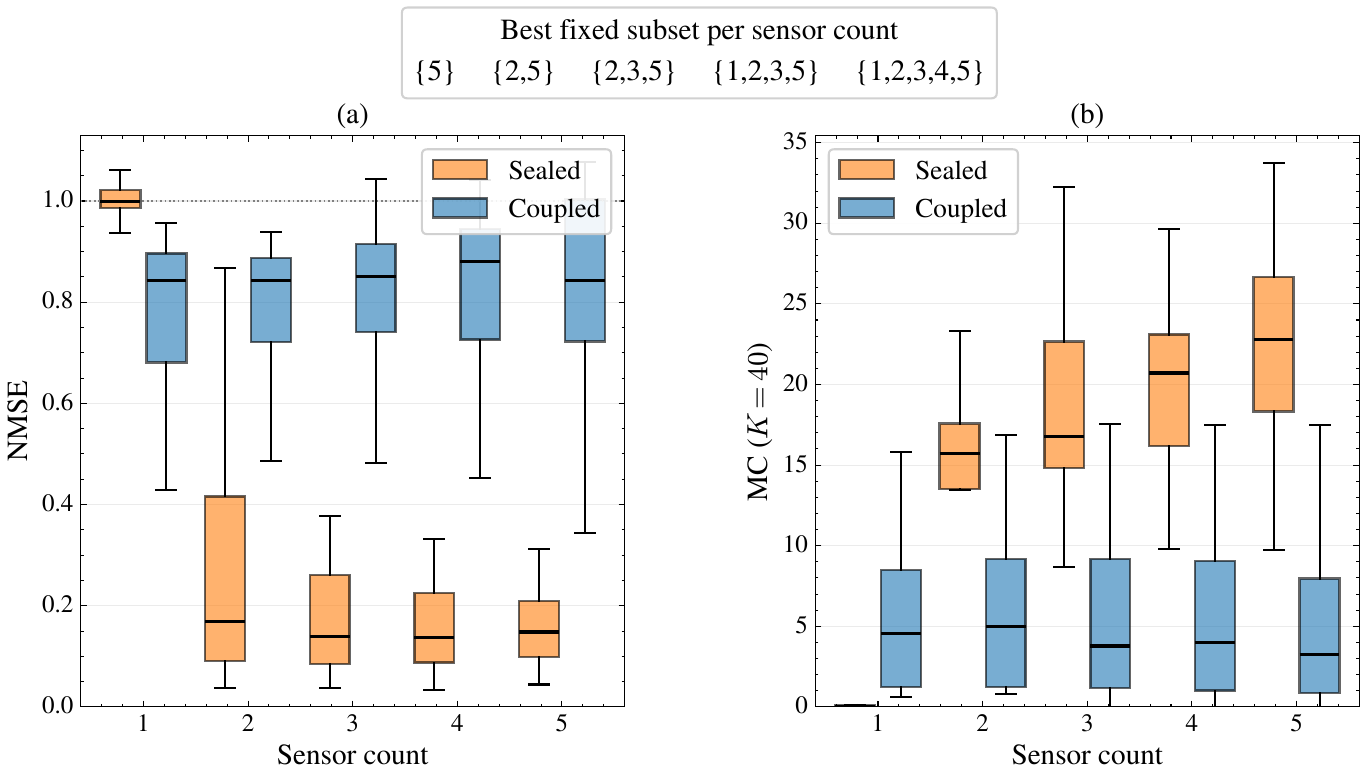}
\caption{Instrumented-sensor count trade-off for the best fixed subset at each $m$ (labelled on the x-axis). Boxes: IQR; centre line: median; whiskers: $1.5{\times}$IQR. (a)~NMSE. (b)~MC ($K{=}40$).}
\label{fig:ablation}
\end{center}
\end{figure}

The optimal subsets follow a spatially interpretable sequence: \{P5\}, \{P2,\,P5\}, \{P2,\,P3,\,P5\}, \{P1,\,P2,\,P3,\,P5\}, and \{P1,\ldots,P5\}, shown by pouch number on the x-axis of Fig.~\ref{fig:ablation}. Pouch P5, located at the base, provides the most informative standalone measurement; P2 introduces distal complementarity at $m{=}2$; and the tip sensor P1 contributes only once broader spatial coverage has been established.

As shown in Fig.~\ref{fig:ablation}(a), a single sensor is insufficient regardless of topology. The critical transition occurs at $m{=}2$: median NMSE in the sealed topology falls from $0.985$ to $0.169$, capturing 97.6\% of the total improvement toward the five-sensor baseline, while the IQR narrows substantially. Beyond two sensors, additional pouches yield diminishing returns (0.139 at m=3, 0.138 at m=4, 0.148 at m=5), with the slight increase at $m{=}5$ indicating that pouch 4 adds redundancy rather than new diversity. Sensor count has no analogous effect in the coupled topology, which remains flat at $0.842$--$0.881$ across all $m$, confirming that the shared manifold is the binding constraint. Fig.~\ref{fig:ablation}(b) mirrors this: sealed MC jumps from near zero at $m{=}1$ to $15.7$ at $m{=}2$ and grows to $22.8$ at $m{=}5$; coupled MC stays low and variable ($3.3$--$5.0$ median, wide IQR) at all counts.

\section{Discussion}
The dominance of topology over stiffness and sensor count suggests that pneumatic interconnection is a structural design decision, not a tuning parameter: once the manifold is shared, no additional sensing or excitation can recover the lost spatial diversity. In reservoir computing, coupling between internal nodes is generally regarded as a source of computational richness \citep{lukosevicius2009rc,tanaka2019prc}, and the shared manifold is exactly such a physical coupling between pouches, so air exchange could plausibly have enriched the reservoir dynamics. Instead, at the tested time scales, manifold equalization acts as spatial averaging that collapses the pouches onto a single global mode (PC1 $\approx 99\%$).

The asymmetry between $P_0$ and $p_{\max}$ has a practical implication: stiffness should be treated as a co-design variable alongside topology, whereas actuation range can be chosen freely for task reasons without affecting reservoir quality. The rapid sensor saturation in the sealed topology suggests that the arm's deformation field is low-dimensional despite having five pouches; strategic placement guided by deformation coverage matters more than sensor count.

Three limitations bound these guidelines. First, all excitations are slow ($f = 0.1$~Hz). At higher frequencies, pneumatic transmission low-pass filters both topologies, and finite manifold flow resistance would make the coupled pouches transiently more independent; we expect the sealed advantage to persist wherever equalization remains fast relative to the motion, but the crossover frequency remains to be characterized. Second, the ridge readout is deliberately linear; a nonlinear readout (e.g., a shallow neural network) might extract more from the coupled signals, although with $\bar{r} = 0.992$ and PC1 $\approx 99\%$ little independent information appears to survive at the sensor level for any readout to exploit. Third, the chronological 70/30 split guards against within-trial leakage but not long-term drift; the topology ranking was consistent across all 36 trials, but stability under fabric hysteresis, wear, and temperature over long operational durations remains to be quantified.

\section{Conclusion}
This work examined pneumatic reservoir sensing as a three-variable design problem. Topology determines whether local pressure histories survive; robot stiffness determines how quickly those histories collapse into redundancy; and instrumented-sensor count determines how much surviving diversity reaches the estimator. The sealed topology produces the most informative multi-sensor state; the coupled manifold remains attractive only for sensitivity-critical or plumbing-constrained designs; increasing $P_0$ pushes the robot toward redundant sensing; and two to three well-placed sensors capture nearly all useful benefit in the sealed design. 

Future work will extend the benchmark beyond periodic excitation and same-time estimation: intermediate topologies, aperiodic and broadband excitation (e.g., multisine and pseudorandom sequences) with complex task-driven motion, higher actuation frequencies, nonlinear readouts, and closed-loop trajectory control using the PRC-based controllers.

\section*{DECLARATION OF GENERATIVE AI TECHNOLOGIES IN THE WRITING PROCESS}
The authors used ChatGPT only for language editing and manuscript structuring; all technical content, interpretation, and final wording were reviewed and approved by the authors, who take full responsibility for the paper.

\bibliography{ifacconf}

@article{tanaka2019prc,
   title={Recent advances in physical reservoir computing: A review},
   volume={115},
   ISSN={0893-6080},
   DOI={10.1016/j.neunet.2019.03.005},
   journal={Neural Networks},
   publisher={Elsevier BV},
   author={Tanaka, Gouhei and Yamane, Toshiyuki and Héroux, Jean Benoit and Nakane, Ryosho and Kanazawa, Naoki and Takeda, Seiji and Numata, Hidetoshi and Nakano, Daiju and Hirose, Akira},
   year={2019},
   month = "July", pages={100–123} }

@techreport{jaeger2001echo,
  title       = {The ``echo state'' approach to analysing and training recurrent neural networks},
  author      = {Jaeger, Herbert},
  institution = {German National Research Center for Information Technology (GMD)},
  number      = {148},
  year        = {2001}
}

@article{maass2002lsm,
  title     = {Real-time computing without stable states: A new framework for neural computation based on perturbations},
  author    = {Maass, Wolfgang and Natschl{\"a}ger, Thomas and Markram, Henry},
  journal   = {Neural Computation},
  volume    = {14},
  number    = {11},
  pages     = {2531--2560},
  year      = {2002},
  month     = {nov},
  publisher = {MIT Press},
  doi       = {10.1162/089976602760407955},
  pmid      = {12433288}
}

@article{lukosevicius2009rc,
title = {Reservoir computing approaches to recurrent neural network training},
journal = {Computer Science Review},
volume = {3},
number = {3},
pages = {127-149},
year = {2009},
issn = {1574-0137},
author = {Mantas Lukoševičius and Herbert Jaeger}
}

@article{nakajima2014softbody,
    author  = {Nakajima, K. and Li, T. and Hauser, H. and Pfeifer, R.},
    title   = {Exploiting short-term memory in soft body dynamics as a computational resource},
    journal = {Journal of The Royal Society Interface},
    volume  = {11},
    number  = {100},
    pages   = {20140437},
    year    = {2014},
    doi     = {10.1098/rsif.2014.0437},
}

@article{rus2015soft,
  author  = {Rus, D. and Tolley, M.},
  title   = {Design, fabrication and control of soft robots},
  journal = {Nature},
  year    = {2015},
  volume  = {521},
  pages   = {467--475},
  doi     = {10.1038/nature14543}
}

@article{wang2025ais_prc_fabric,
author = {Wang, Jun and Qiao, Zhi and Zhang, Wenlong and Li, Suyi},
title = {Proprioceptive and Exteroceptive Information Perception in a Fabric Soft Robotic Arm via Physical Reservoir Computing with Minimal Training Data},
journal = {Advanced Intelligent Systems},
volume = {7},
number = {4},
pages = {2400534},
eprint = {https://advanced.onlinelibrary.wiley.com/doi/pdf/10.1002/aisy.202400534},
year = {2025}
}

@article{shen2025fprc_hysteresis,
  author={Shen, Junyi and Miyazaki, Tetsuro and Kawashima, Kenji},
  journal={IEEE Robotics and Automation Letters}, 
  title={Control Pneumatic Soft Bending Actuator With Feedforward Hysteresis Compensation by Pneumatic Physical Reservoir Computing}, 
  year={2025},
  volume={10},
  number={2},
  pages={1664-1671},
  doi={10.1109/LRA.2024.3523229}}

@article{shen2025gainscheduled_stateest,
  author={Shen, Junyi and Miyazaki, Tetsuro and Kawashima, Kenji},
  journal={IEEE Robotics and Automation Letters}, 
  title={Controlling Pneumatic Bending Actuator With Gain-Scheduled Feedforward and Physical Reservoir Computing State Estimation}, 
  year={2025},
  volume={10},
  number={3},
  pages={2120-2127},
  doi={10.1109/LRA.2025.3528661}}

@incollection{lukosevicius2012practical,
  title={A Practical Guide to Applying Echo State Networks},
  author={Luko{\v{s}}evi{\v{c}}ius, Mantas},
  booktitle={Neural Networks: Tricks of the Trade},
  editor={Montavon, Grégoire and Orr, Geneviève B. and Müller, Klaus-Robert},
  publisher={Springer},
  address={Berlin, Heidelberg},
  year={2012},
  pages={659--686},
  doi={10.1007/978-3-642-35289-8\_36}
}

@article{eder2018morph_control,
  author    = {M. Eder and F. Hisch and H. Hauser},
  title     = {Morphological computation-based control of a modular, pneumatically driven, soft robotic arm},
  journal   = {Advanced Robotics},
  volume    = {32},
  number    = {7},
  pages     = {375--385},
  year      = {2018},
  publisher = {Taylor \& Francis},
  doi       = {10.1080/01691864.2017.1402703},
  eprint    = {https://doi.org/10.1080/01691864.2017.1402703}
}

@article{dambre2012ipc,
  author  = {Dambre, Joni and Verstraeten, David and Schrauwen, Benjamin and Massar, Serge},
  title   = {Information Processing Capacity of Dynamical Systems},
  journal = {Scientific Reports},
  volume  = {2},
  pages   = {514},
  year    = {2012},
  doi     = {10.1038/srep00514},
}

@article{lin1989ccc,
 ISSN = {0006341X, 15410420},
 author = {Lawrence I-Kuei Lin},
 journal = {Biometrics},
 number = {1},
 pages = {255--268},
 publisher = {[Wiley, International Biometric Society]},
 title = {A Concordance Correlation Coefficient to Evaluate Reproducibility},
 urldate = {2026-04-18},
 volume = {45},
 year = {1989}
}

@article{nakajima2015infoproc,
  author  = {Nakajima, Kohei and Hauser, Helmut and Li, Tao and Pfeifer, Rolf},
  title   = {Information processing via physical soft body},
  journal = {Scientific Reports},
  volume  = {5},
  pages   = {10487},
  year    = {2015},
  doi     = {10.1038/srep10487},
}

@article{nguyen2020fabric,
  author  = {Nguyen, P. H. and Zhang, W.},
  title   = {Design and Computational Modeling of Fabric Soft Pneumatic Actuators for Wearable Assistive Devices},
  journal = {Scientific Reports},
  volume  = {10},
  pages   = {9638},
  year    = {2020},
  doi     = {10.1038/s41598-020-65003-2},
  
}

@article{qiao2024nonlinear,
  title={Nonlinear Disturbance Observer with Sliding Mode Control for a Fabric Soft Robotic Arm},
  author={Qiao, Zhi and Tao, Weijia and Zhang, Wenlong},
  journal={IFAC-PapersOnLine},
  volume={58},
  number={28},
  pages={516--521},
  year={2024},
  publisher={Elsevier}
}

\end{document}